\documentclass[10pt,journal,compsoc]{IEEEtran}

\ifCLASSOPTIONcompsoc
  \usepackage[nocompress]{cite}
\else
  \usepackage{cite}
\fi

\ifCLASSINFOpdf
\else
\fi

\usepackage{graphicx}

\usepackage{amsmath,amsfonts,bm}

\def\eqref#1{equation~\ref{#1}}

\def\1{\bm{1}}

\DeclareMathAlphabet{\mathsfit}{\encodingdefault}{\sfdefault}{m}{sl}
\SetMathAlphabet{\mathsfit}{bold}{\encodingdefault}{\sfdefault}{bx}{n}

\usepackage[export]{adjustbox}%
\usepackage{tabularx}
\usepackage{array}
\usepackage{multirow} 
\usepackage{tabularx}
\usepackage{booktabs}  
\usepackage{subcaption} 

\usepackage{color} 
\usepackage[pagebackref=true,breaklinks=true,letterpaper=true,colorlinks,bookmarks=false]{hyperref}
\usepackage[dvipsnames]{xcolor}
\usepackage[misc]{ifsym}

\usepackage{ulem}
\usepackage{algorithm}
\usepackage{algpseudocode} 
\usepackage{breqn} 
\usepackage[inline]{enumitem}

\usepackage{pifont}
\newcommand{\cmark}{\ding{51}}%
\newcommand{\xmark}{\ding{55}}%

\iftrue

\newcommand{\SQ}[1]{{\color{black}{#1}}}
\newcommand{\YX}[1]{{\color{red}{\bf YX:} #1}}
\newcommand{\alertJW}[1]{{\color{magenta}{\bf JW:} #1}}
\else
\newcommand{\SQ}[1]{}
\newcommand{\alertJW}[1]{}
\newcommand{\YX}[1]{}
\newcommand{\CS}[1]{}
\newcommand{\LH}[1]{}
\newcommand{\FK}[1]{}

\usepackage{hyperref}
\usepackage{url}
\usepackage{microtype}      
\usepackage{xcolor}         
\usepackage{enumitem}
\usepackage{times}
\usepackage{epsfig}
\usepackage{graphicx}
\usepackage{amsmath}
\usepackage{amssymb}
\usepackage{comment}
\usepackage{bm}

\usepackage[utf8]{inputenc}
\usepackage[T1]{fontenc}

\usepackage{booktabs} 

\usepackage{threeparttable}
\usepackage{times}
\usepackage{epsfig}

\usepackage{natbib} 
\usepackage{caption}
\usepackage{color}

\newcommand{\SQ}[1]{{\color{black}{#1}}}
\newcommand{\YX}[1]{{\color{red}{\bf YX:} #1}}

\usepackage{colortbl}
\usepackage{amsthm}
\usepackage{subfigure} 
\usepackage{algorithm}
\usepackage{multirow}
\usepackage{dsfont}
\usepackage{color}
\usepackage{wrapfig}
\usepackage{pifont}
\newcommand{\cmark}{\ding{51}}%
\newcommand{\xmark}{\ding{55}}%
\fi

\begin{document}
%
\title{OneRing: A Simple Method for Source-free Open-partial Domain Adaptation}

\author{Shiqi Yang,
        Yaxing Wang$^{(\textrm{\Letter})}$, Kai Wang,
        Shangling Jui, \and Joost van de Weijer
\IEEEcompsocitemizethanks{\IEEEcompsocthanksitem S. Yang, Kai Wang and J. van de Weijer are with the Computer Vision Center, Universitat Aut\`onoma de Barcelona, Barcelona 08193, Spain.\protect\\
E-mail: \{syang,kwang,joost\}@cvc.uab.es.
\IEEEcompsocthanksitem Y. Wang, College of Computer Science, Nankai University,  China. E-mail:yaxing@nankai.edu.cn.
\IEEEcompsocthanksitem S. Jui  is  with Huawei Kirin Solution, Shanghai, China. E-mail:jui.shangling@huawei.com.
}
\thanks{Manuscript received April 19, 2005; revised August 26, 2015.}}
%
%

\markboth{Journal of \LaTeX\ Class Files,~Vol.~14, No.~8, August~2015}%
{Shell \MakeLowercase{\textit{et al.}}: Bare Demo of IEEEtran.cls for Computer Society Journals}
%



\IEEEtitleabstractindextext{%
\begin{abstract}
In this paper, we investigate \textit{Source-free \SQ{Open-partial} Domain Adaptation} (SF-OPDA), which addresses the situation where there exist both domain and category shifts between source and target domains. Under the SF-OPDA setting, which aims to address data privacy concerns, the model cannot access source data anymore during target adaptation. We propose a novel training scheme to learn a ($n$+1)-way classifier to predict the $n$ source classes and the unknown class, where samples of only known source categories are available for training. Furthermore, for target adaptation, we simply adopt a weighted entropy minimization to adapt the source pretrained model to the unlabeled target domain without source data. In experiments, we show our simple method surpasses current OPDA approaches which demand source data during adaptation. When augmented with a closed-set domain adaptation approach during target adaptation, our source-free method further outperforms the current state-of-the-art OPDA method by 2.5\%, 7.2\% and 13\% on Office-31, Office-Home and VisDA respectively.
\end{abstract}

}

\maketitle

\IEEEdisplaynontitleabstractindextext

%
\IEEEpeerreviewmaketitle

\IEEEraisesectionheading{\section{Introduction}\label{sec:intro}}

%
%
%
%


\IEEEPARstart{M}{odern} deep learning models excel at closed-set recognition tasks across various computer vision application areas. However, there are several inevitable obstacles on the path to deploying these methods to challenging real-world environments. As there may be 1) some unseen categories, or 2) a distributional shift between training and testing data. The first problem is usually defined as \textit{open-set recognition} ({OSR})~\cite{Chen_2020_ECCV,Ge2017Generative,Neal_2018_ECCV,Sun_2020_CVPR,zhang2020hybrid,Shu2020podn,vaze2022openset} where the model should be able to distinguish samples from seen or unseen categories. The second problem is mostly investigated in \textit{domain generalization} ({DG})~\cite{shi2021gradient,robey2021model,vedantam2021empirical,gulrajani2020search,wang2021learning} and \textit{domain adaptation} ({DA})~\cite{long2018transferable,long2015learning,long2016unsupervised,tzeng2017adversarial,zhang2019domain,cicek2019unsupervised,liang2021domain,deng2019cluster,tang2020unsupervised,cui2020towards}. DG aims to tackle the domain shift problem in the absence of target domains, while DA seeks to transfer knowledge from labeled source domains to unlabeled target domains while training on both labeled source and unlabeled target data. 
In recent years, several works introduce open-set recognition into DG and DA, which are formalized as \textit{open domain generalization} ({ODG})~\cite{shu2021open,zhu2022crossmatch}, \textit{open-set domain adaptation} ({OSDA})~\cite{saito2018open,bucci2020effectiveness,liu2019separate,pan2020exploring,jing2021towards,feng2019attract,feng2021open} and \textit{universal domain adaptation} ({UNDA})/\textit{open-partial domain adaptation} (OPDA)~\cite{fu2020learning,li2021domain,you2019universal,saito2020universal,saito2021ovanet,liang2021umad}, respectively.

\begin{table*}[tpb]
\caption{Related setting: $\mathcal{C}_s$ and $\mathcal{C}_t$ denote label set of source and target domain (for evaluation), $\mathcal{P}_s$ and $\mathcal{P}_t$ denote source and target distribution, transductive means model can be trained on target data.\vspace{0mm}}
      \label{tab:settings}
      \centering
      \addtolength{\tabcolsep}{6pt}
\scalebox{0.99}
{
      \begin{tabular}{c|c|c|c}
      \hline
           \textbf{Task}& {$\mathcal{C}_s=\mathcal{C}_t$} & {{$\mathcal{P}_s = \mathcal{P}_t$}}& {\textbf{Transductive}}
           \\
           \hline
           {\textit{Open-set Recognition} (OSR)}  &\xmark  &\cmark &\xmark  \\
           \hline
           {\textit{Domain Generalization} (DG)} &\cmark  &\xmark &\xmark   \\
          {\textit{Open Domain Generalization} (ODG)} &\xmark  &\xmark &\xmark  \\
           \hline
            {\textit{Domain Adaptation} (DA)}&\cmark  &\xmark &\cmark    \\
           {\textit{Open-partial Domain adaptation (OPDA)}}&\xmark  &\xmark &\cmark  \\
           \hline
      \end{tabular}}
\end{table*}

The various settings described above are summarized in Table.~\ref{tab:settings}. Usually one method tailored for a specific setting in Table.~\ref{tab:settings} does not work well under a different setting. Most existing works in \textit{Open-set Recognition} are computationally demanding, either requiring the generation of unknown categories~\cite{Neal_2018_ECCV} or conducting additional learning~\cite{kong2021,Sun_2020_CVPR,Chen_2020_ECCV}. Additionally, those methods are likely to suffer from performance degradation if test data are from different distributions.
The recent CrossMatch~\cite{zhu2022crossmatch} tackles the \textit{Open-set Single Domain Generalization} problem. It proposes to use multiple open class detectors which are put on top of existing single domain generalization methods, and it achieves good results at the expense of introducing multiple open-set detectors and auxiliary unknown sample generation. For \textit{open-partial domain adaptation}, most works are based on an explicitly designed unknown-sample rejection module, which typically requires various hyperparameters. More importantly, those \textit{OPDA}/\textit{UNDA} methods all require access to source data during target adaptation, which is infeasible when having data privacy issues and when deployed on devices of low computational capacity.

In this paper, we investigate the challenging source-free open-partial domain adaptation task\footnote{\textbf{Most recent papers refer to open-partial domain adaptation as universal DA~\cite{changunified,saito2021ovanet,kundusubsidiary}, in this paper we adopt open-partial DA to avoid any potential misunderstanding.}}, where there is no access to source data during adaptation. Thus, a question arises, how to build a model training from only known categories aiming to learn to distinguish
samples of known and unknown categories? 
Since we have no access to unknown class data, we can only use the known class data to train this classifier. We hypothesize that the closest (most similar) class to any known class {can be} an unknown class. Given the open-endedness of the unknown class, this is a reasonable assumption. This hypothesis allows us to train the classifier, enforcing the most probable class to be the ground truth class, and the runner-up class to be the background class for all source data.
This is achieved by introducing an extra category in the classifier which represents the unknown classes, during training on samples of known categories (yielding a ($n$ + 1)-way classifier where $n$ is the number of known classes), the classifier is expected to output the largest score for the ground truth class, and the second-largest score for the unknown class. By this way, the model can learn to reject samples of unknown categories by only training with known classes.

Furthermore, 
our source model with strong capacity to distinguish unknown categories can be easily adapted to the target domain without access to source data under the challenging \textit{source-free open-partial domain adaptation} setting, where both source and target domains have their private classes. We propose to simply use a weighted entropy minimization to achieve the adaptation.

We summarize our contributions as below:
\begin{itemize}[leftmargin=*]
    \item We propose a simple method called \textit{OneRing}, which excels at recognizing data of unknown classes after source training, 
    thus it can be easily adapted the source model to target domain by using weighted entropy minimization under \textit{source-free open-partial domain adaptation} setting (SF-OPDA).
    \item In experiments, we show our method is on par with or outperforms current state-of-the-art approaches on several benchmarks, which proves the efficacy and generalization ability of our method. Augmented with a closed-set DA approach, our source-free method surpasses current open-partial domain adaptation methods by a significant margin.
\end{itemize}

\section{Related Works}
\textbf{Open-set Recognition.}
\textit{Open-set recognition} (OSR) aims to recognize samples of unknown categories which do not exist in the training set. Several recent methods in OSR do not utilize extra data for training. OpenHybrid~\cite{zhang2020hybrid} introduces a flow-based density estimation module, and ARPL~\cite{Chen_2020_ECCV,chen2021adversarial} proposes to learn a reciprocal point per category, which is intuitively regarded as the farthest point from the corresponding feature group. More recently \cite{vaze2022openset} shows that actually OSR performance is enhanced when improving the model performance on the training set, for example by using improved data augmentation and other training tricks. In this paper, we propose a simple model training directly with two cross entropy losses without either auxiliary data or an extra learning process. {Our proposed OneRing classifier shares similarity with Proser~\cite{zhou2021learning}, which aims to assign the second-largest logit to the unknown classes. However, Proser is much more complex compared to ours: it first trains a good $|C_s|$-way closed-set classifier and then augment this classifier to $|C_s|+C$-way, and retrain; Further, it needs to synthesize novel samples for training the $|C_s|+C$-way classifier; And they also need to calibrate the output of the dummy classifier over the extra validation set by ensuring 95\% of validation data are recognized as known. While in this paper, we directly train the $|C_s|+1$-way classifier with a simple objective; Another main difference is that they only address open-set recognition, while in our paper we also consider the domain shift, \textit{i.e.}, the challenging source-free open-partial domain adaptation. 


\textbf{Domain Adaptation.}
Early methods to tackle {\textit{domain adaptation} (DA)} conduct feature alignment~\cite{long2015learning,sun2016return,tzeng2014deep} to eliminate the domain shift. DANN~\cite{ganin2016domain}, CDAN~\cite{long2018conditional} and DIRT-T~\cite{shu2018dirt} further resort to adversarial training to learn domain invariable features. Similarly, \cite{lee2019sliced,lu2020stochastic,saito2018maximum} are based on multiple classifier discrepancy to achieve alignment between domains. Other methods like SRDC~\cite{tang2020unsupervised}, CST~\cite{liu2021cycle} address domain shift from the perspective of either clustering or improved pseudo labeling. And there are also methods considering category shift source and target domains. They can be grouped into \textit{partial-set DA}~\cite{cao2018partial,cao2019learning,liang2020balanced}, \textit{open-set DA}~\cite{panareda2017open,saito2018open,liu2019separate,bucci2020effectiveness} and \textit{universal DA}~\cite{you2019universal,li2021domain,saito2021ovanet,fu2020learning,saito2020universal} depending on the intersection degree of source and target label space. {OVANet~\cite{saito2021ovanet} is a universal DA method. It trains extra $n$ binary classifiers with hard negative classifier sampling to reject unknown samples, OVANet needs to check the normal classifier head and the corresponding binary classifier for the final prediction. While in this paper, we simply train a $n+1$-way classifier with normal cross entropy, and the final prediction is directly provided by the classifier.} 

\textbf{Source-free Domain Adaptation.}
Recently, several works address {\textit{source-free domain adaptation} (SFDA)}, where a source pretrained model is adapted to target without source data. SHOT~\cite{liang2020we} proposes to use mutual information maximization along with pseudo labeling. BAIT~\cite{yang2020unsupervised} adapts MCD~\cite{saito2018maximum} to source-free setting. 3C-GAN~\cite{li2020model} resorts to fake target-style images generation. HCL~\cite{huang2021model} conducts Instance Discrimination~\cite{wu2018unsupervised} over different historical models to cluster features, with the companion of pseudo labeling. $A^2$Net~\cite{xia2021adaptive} learns extra classifier specifically for the target domain and introduce a category-wise matching module for feature clustering. G-SFDA~\cite{yang2021generalized} and NRC~\cite{yang2021exploiting} are all based on neighborhood clustering through local prediction consistency. AaD~\cite{yang2022attracting} further treats SFDA as a typical unsupervised clustering problem and proposes to optimize an upper bound of a clustering objective.
Beyond closed-set DA, FS~\cite{kundu2020towards} and USFDA~\cite{kundu2020universal}, which are for \textit{source-free open-set and open-partial DA} respectively. However, they both synthesize extra training samples of unknown categories, which help to detect the open classes. OSHT~\cite{feng2021open} tackles source-free open-set DA, which adopts pseudo labeling for adaptation and entropy-based metric to reject open classes.
In this paper, we show that our source pretrained model can be adapted to the target domain easily by simply minimizing entropy to achieve \textit{source-free open-partial DA}.





\section{Method}
\subsection{Preliminary}
Data samples are divided into two domains: the labeled source domain with $N_s$ samples $\mathcal{D}_s=\{(x_i^s,y_i^s)\}_{i=1}^{N_s}$ on which the model will be first trained, and the unlabeled target domain with $N_t$ samples $\mathcal{D}_t=\{x_i^t\}_{i=1}^{N_t}$. $\mathcal{D}_t$ is used for evaluation. We denote $\mathcal{C}_s$ and $\mathcal{C}_t$ as the label set of the source and target domain, and {$\mathcal{P}_s$ and $\mathcal{P}_t$ as the distribution of source and target data}, respectively. 
In this paper, we study the challenging
\textbf{\textit{Source-free open-partial domain adaptation} (SF-OPDA)} task, where the source pretrained model has to adapt to the target domain without access to any source data and both domains have private categories {($\mathcal{C}_s \cap \mathcal{C}_t \neq\text{ any of }\{\emptyset, \mathcal{C}_s, \mathcal{C}_t\}, \mathcal{P}_s\neq\mathcal{P}_t$).} 


\begin{figure*}[tpb]
    \centering
    \includegraphics[width=0.99\linewidth]{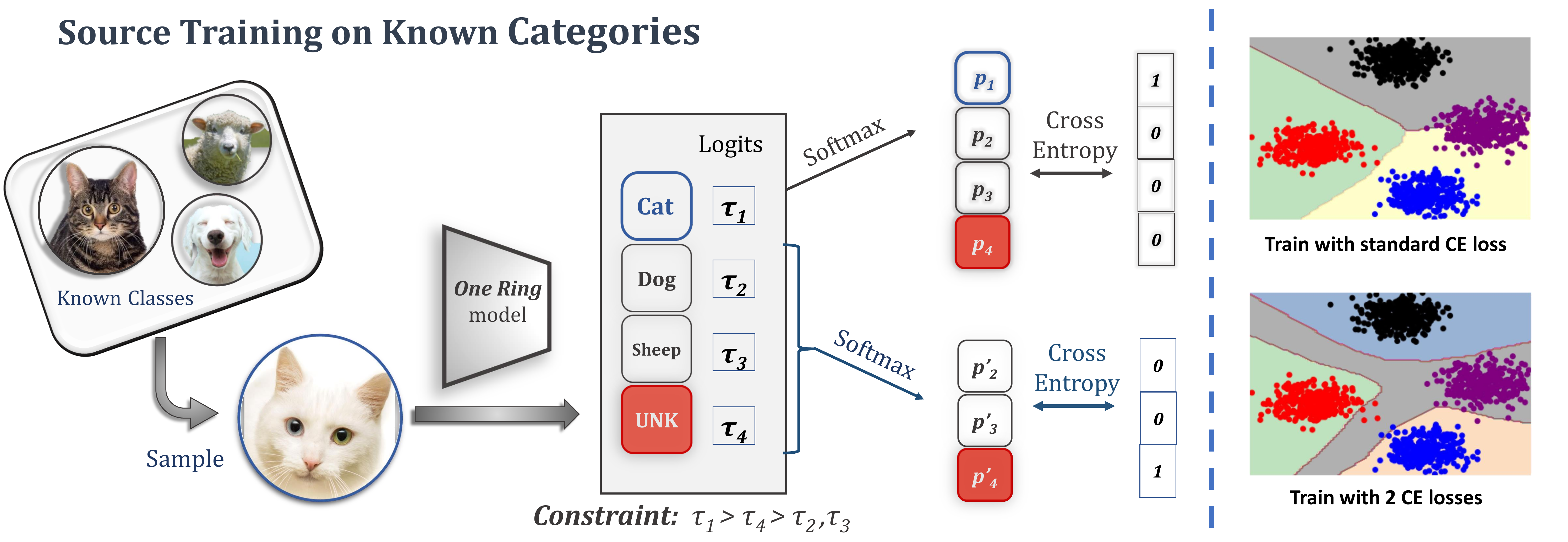}
    \caption{(\textbf{Left}) Illustration of training \textit{OneRing} model on source data with only known categories. (\textbf{Right}) Toy Example, the decision boundaries and prediction regions (\textit{colorized randomly}) after training on 3 known classes with ($3 + 1$)-way classifier. Purple points are from unknown category.\vspace{0mm}}\label{fig:onering-s_toy}
\end{figure*}

\subsection{Source Training: One Ring to Find Unknown Categories}
The first stage is to train a model on the labeled source domain which has $|\mathcal{C}_s|$ categories. We expect the resulting model to have the ability to detect unknown categories which do not exist in the source data. To achieve this, we build a classifier head as a ($|\mathcal{C}_s| \bm{+ 1}$)-way classifier, where the additional dimension aims to distinguish unknown categories. 
Then the following problem arises: how to train a ($|\mathcal{C}_s| + 1$)-way classifier without any sample from the unknown category? Note, if only training with the normal cross entropy (CE) loss on the source data, the model cannot directly give prediction to unknown categories.




As mentioned in Sec.~\ref{sec:intro}, we hypothesize that any non-ground-truth category could be regarded as an unknown category. This hypothesis gives us a feasible solution to train an open-set classifier without actually accessing open classes. Specifically,
we propose to use a simple variant of cross entropy loss with only samples of known categories to train the ($|\mathcal{C}_s| + 1$)-way classifier, which has 2 properties: 1) The largest output logit of the source samples corresponds to the ground truth class and 2) The second-largest output logit of source samples will be the unknown class (($|\mathcal{C}_s| + 1$)-th class in classifier). This way, the model is expected to detect samples of unknown categories even without training on them. The proposed objective to achieve it is formalized as follows: 
\begin{equation}\label{eq:ring_src}
    \mathcal{L}_{source}=\mathbb{E}_{x_i\sim \mathcal{D}_s}[\mathcal{L}_{\textit{ce}}(p(x_i),y_i)+\mathcal{L}_{\textit{ce}}(\hat{p}(x_i),\hat{y_i})]
\end{equation}
where $p(x_i)=g(f(x_i))\in \mathbb{R}^{|C_s|+1}$ is the output vector of the ($|\mathcal{C}_s| + 1$)-way classifier, while $\hat{p}(x_i)\in \mathbb{R}^{|C_s|}$ is the output vector removing the dimension corresponding to the ground truth class, and $\hat{y}_i \in \mathbb{R}^{|C_s|}$ is a one-hot label with unknown class as ground truth label. As illustrated in Fig.~\ref{fig:onering-s_toy} (\textit{right}), if we have a sample $x_i$ belonging to the \textbf{\textit{first}} class, the first CE loss in Eq.~\ref{eq:ring_src} is the typical CE loss on $p(x_i)$ with ground truth label, $\hat{p}(x_i)$ is produced by removing the \textbf{\textit{first}} dimension and the second CE loss is applied on $\hat{p}(x_i)$ with unknown (last) category as label. 

We adopt a toy example to illustrate it. As shown in upper part of Fig.~\ref{fig:onering-s_toy} (\textit{right}), we generate isotropic Gaussian blobs with 4 categories, where the last one is treated as the unknown category (in \textit{Purple}) and others as known classes (thus $|\mathcal{C}_s|=3$). We first train the ($|\mathcal{C}_s| + 1$)-way classifier with the normal cross entropy loss on samples of known categories, and then evaluate it on all classes. The upper part of Fig.~\ref{fig:onering-s_toy} (\textit{right}) shows that the samples of the hypothetical unknown category (\textit{Purple}) are misclassified as there are only 3 prediction regions for 3 known categories.
As shown in the lower part of Fig.~\ref{fig:onering-s_toy} (\textit{right}) that there are 4 prediction regions (3 known + 1 unknown categories), after training on 2 CE losses the classifier can detect samples of unknown category which is unseen before. 

An intuitive understanding of the proposed method is that we can split the ($|\mathcal{C}_s| + 1$)-way classification into 2 levels: 1) if we check the prediction $p(x_i)$ we would say $x_i$ has to belong to category $y_i$; 2) if we check the prediction $\hat{p}(x_i)$ we would say that $x_i$ is impossible to belong to all other categories except the potential unknown categories. Since in Eq.~\ref{eq:ring_src} the output score of unknown category (last dimension) will always rule other non-ground-truth categories, we call the last dimension of the classifier head as \textit{OneRing} dimension and our model as \textit{OneRing}. 

\subsection{Target Adaptation: One Ring to Bind All Categories without the Source}
Our source-pretrained \textit{OneRing} model is empowered with the ability to recognition unknown classes in the target domain. We further posit that it can easily be adapted to target domains where domain shift and unknown categories exist. The key part is to rectify the wrong predictions due to the domain shift. We propose to simply use entropy minimization, which is widely used in DA~\cite{shu2018dirt,long2018conditional,liang2020we,saito2020universal,saito2021ovanet}, to achieve adaptation with only a slight but indispensable modification:
\begin{equation}\label{eq:ring_tar}
    \mathcal{L}_{target}=\frac{bs}{\hat{n}_{k_{all}}}\mathbb{E}_{\bar{y_i} \in \mathcal{C}_s} \mathcal{L}_{\textit{ent}}(p(x_i))+\frac{bs}{\hat{n}_{u_{all}}}\mathbb{E}_{\bar{y_i} \in \mathcal{C}_u} \mathcal{L}_{\textit{ent}}(p(x_i)) 
\end{equation}
which is computed in the \textbf{mini-batch} ($bs$ denotes batch size), and $\bar{y_i}$ is the predicted label, {$\hat{n}_{k_{all}}$ is the number of samples in the \textit{whole dataset} which are predicted as \textit{known} category $\mathcal{C}_s$, $\hat{n}_{u_{all}}$ is the number of those predicted as \textit{unknown} category $\mathcal{C}_u$ also in the \textit{whole dataset}. 
Here $\frac{bs}{\hat{n}_{k_{all}}}=\frac{N_t}{\hat{n}_{k_{all}}} \times \frac{bs}{N_t}$ (similar for $\frac{bs}{\hat{n}_{u_{all}}}$), where $N_t=\hat{n}_{k_{all}}+\hat{n}_{u_{all}}$ and $\frac{N_t}{\hat{n}_{k_{all}}}$ is the reciprocal of the known/unknown category ratio (prior information according to the predictions). The reason to deploy these weights is to balance the two entropy terms, and $\frac{bs}{N_t}$ is a scale factor. \footnote{Instead of using the predictions over the whole dataset to compute known-unknown ratio, we can also use prediction of current mini-batch for approximation (thus $N_t$ will be replaced by $bs$, and similar for $\hat{n}_{u_{all}}$ and $\hat{n}_{k_{all}}$), in the experiment we empirically found these two different estimation manners lead to almost the same results.}}
With this simple objective, the source model can be adapted to the target domain under domain and category shift efficiently.

\noindent\textbf{Augmented with Attracting-and-Dispersing.}
Since our \textit{OneRing} method can equip models to efficiently detect unknown classes, it can be used as a baseline to be combined with methods in closed-set source-free DA. Here we integrate our method with a simple state-of-the-art SFDA method Attracting-and-Dispersing (AaD)~\cite{yang2022attracting}, note AaD can not directly tackle the open-partial domain adaptation setting. AaD has an objective with only 2 dot product terms: $\mathcal{L}_{dis}$ for discriminability and $\mathcal{L}_{div}$ for diversity, more details can be found in AaD paper. The resulting objective is:
\begin{equation}\label{eq:final}
\begin{aligned}
    \mathcal{L}_{target+}=\frac{bs}{\hat{n}_{k_{all}}}\mathbb{E}_{\bar{y_i} \in \mathcal{C}_s}[ \mathcal{L}_{\textit{ent}}(p(x_i))+\mathcal{L}_{dis}+\mathcal{L}_{div}]\\ +\frac{bs}{\hat{n}_{u_{all}}}\mathbb{E}_{\bar{y_i} \in \mathcal{C}_u} [\mathcal{L}_{\textit{ent}}(p(x_i))+\mathcal{L}_{dis}]
\end{aligned}
\end{equation}
where we do not deploy the diversity term for samples predicted as an unknown class since there is only one single unknown class. 

\section{Experiments}

Here we provide quantitative results and analyses related to source-free open-partial domain adaptation. 

\subsection{Datasets}


For SF-OPDA, the model is trained on the source domain first, then adapted to the target domain without access to any source data. Here both the source and target domains have their private categories and the target domain has some unknown categories. We evaluate our method on several benchmarks following the same setting as previous work in OPDA~\cite{you2019universal,saito2020universal,saito2021ovanet}: 1) \textbf{Office-31} shares 10 classes with Caltech-256 which will be used as the common categories. Then the next 10 classes in alphabetical order will be source private, and the remaining classes will be target private. 2) \textbf{Office-Home} The first 10 classes in alphabetical order are shared between domains, and the next 5 categories will be source private, and the remaining classes are target private. 3) \textbf{VisDA} (VisDA-C 2017)~\cite{peng2017visda} The 6 classes out of 12 classes will be the shared categories, and source and target domain both have 3 private classes. 4) \textbf{DomainNet}~\cite{peng2019moment} DomainNet is one of the largest domain adaptation benchmarks with around 0.6 million images. Following previous works, we will use 3 domains: Painting (P), Real (R), and Sketch (S). We will use the first 150 classes as shared categories, the next 50 classes are source private and the remaining 145 as target private. The number of source, target and shared categories is described in the title of each table.

\begin{table*}
\begin{minipage}[!h]{0.99\textwidth}
\centering
\addtolength{\tabcolsep}{1pt}
\begin{center}\caption{Accuracy (\%) on \textbf{Office-31} and \textbf{VisDA} dataset using ResNet-50. \textbf{open-partial domain adaptation} where for \textit{Office-31}: $|\mathcal{C}_s|=20$, $|\mathcal{C}_t|=21$, $|\mathcal{C}_s \cap \mathcal{C}_t| =10$; and for \textit{VisDA}: $|\mathcal{C}_s|=9$, $|\mathcal{C}_t|=9$, $|\mathcal{C}_s \cap \mathcal{C}_t| =6$. The second highest H score is underlined. \textbf{SF} indicates whether source-free.\vspace{0mm}}\label{tab:office31_uni}
\scalebox{0.99}{\begin{tabular}{c|c|cccccccccccccc|c} 
\hline
   \multirow{2}{*}{\textbf{Office-31}}  & \multirow{2}{*}{\textbf{SF}} & \multicolumn{2}{c}{A2W} & \multicolumn{2}{c}{D2W} & \multicolumn{2}{c}{W2D} & \multicolumn{2}{c}{A2D} & \multicolumn{2}{c}{D2A} & \multicolumn{2}{c}{W2A} & \multicolumn{2}{c|}{\textbf{Avg}}&\multicolumn{1}{c}{\textbf{VisDA}} \\ 
       &                     & OS  & H                & OS  & H                & OS  & H                & OS  & H                & OS  & H                & OS  & H                & OS  & H        & H         \\ 
\cline{1-17}
OSBP~\cite{saito2018open}   & \xmark                   & 66.1 & 50.2             & 73.6 & 55.5             & 85.6 & 57.2             & 72.9 & 51.1             & 47.4 & 49.8             & 60.5 & 50.2             & 67.7 & 52.3       &27.3       \\
UAN~\cite{you2019universal}    & \xmark                   & 85.6 & 58.6             & 94.8 & 70.6             &\textbf{98.0}   & 71.4             & 86.5 & 59.7             & 85.5 & 60.1             & 85.1 & 60.3             & 89.2 & 63.5  &30.5            \\
ROS~\cite{bucci2020effectiveness}    & \xmark                   & - & 71.3             & - & 94.6             &-   & 95.3             & - & 71.4             & - & 81.0             & - & 81.2             & - & 82.1  &-            \\
CMU~\cite{fu2020learning}    & \xmark                   & 86.7 & 67.3             & \textbf{96.7} & 79.3             & \textbf{98.0}   & 80.4             & 89.1 & 68.1             & 88.4 & 71.4             & 88.6 & 72.2             & 91.1 & 73.1         &34.6     \\
DCC~\cite{li2021domain}    & \xmark                   & \textbf{91.7} & 78.5             & 94.5 & 79.3             & 96.2 & 88.6             & \textbf{93.7} & \textbf{88.5}             & \textbf{90.4} & 70.2             & \textbf{92.0}   & 75.9             & \textbf{93.1} & 80.2       &43.0       \\
\SQ{DANCE}~\cite{saito2020universal} & \xmark                   &  -    & 71.5             &    -  & 91.4            &   -   &87.9             &   -   & 78.6             &    -  &79.9             &  -    & 72.2              &   -   & 80.3 & 4.4             \\ 
OVANet~\cite{saito2021ovanet} & \xmark                   &  -    & 79.4             &    -  & \textbf{95.4}            &   -   & 94.3             &   -   & 85.8             &    -  & 80.1             &  -    & 84.0              &   -   & 86.5 & 53.1              \\ 
\SQ{USFDA}~\cite{kundu2020universal}& \cmark                   &  -&79.8 &-&90.6 &-&81.2 &-&85.5&-&
83.2&-& 88.7&-&
84.8&
-           \\ 
\hline
\textbf{\textit{OneRing}-S}   &    &  69.0        & 67.9 & 92.5             & 90.6 & 96.5            & 89.4 & 81.9              & 74.9 & 64.8             & 74.8   & 69.9             & 78.8   &   79.1           &           79.4      & 35.2
       \\
\textbf{\textit{OneRing}}   &   \cmark                & 78.8 & {83.8}             & 94.7 & 95.2             & 97.5 & {96.0}               & 86.6 & 85.7             & 82.0   & {85.8}             & 81.0   & {84.7}             & 86.8     & \underline{88.5}     & \underline{60.7}        
       \\
\textbf{\textit{OneRing}+}   &   \cmark        &         85.3  &   \textbf{85.4}&     94.0  &   94.2&   97.0&    93.6 &  88.4 &      86.1&   88.9&       \textbf{90.7} &  87.3 &      84.0 &   90.2&      \textbf{89.0} & \textbf{66.1}

       \\
\hline
\end{tabular}}\end{center}
\end{minipage}\vspace{2mm}
\begin{minipage}[!h]{0.99\textwidth}
\centering
\addtolength{\tabcolsep}{2pt}
\begin{center}\caption{\textbf{H-score} (\%) on \textbf{Office-Home} dataset using ResNet-50 as backbone. \textbf{open-partial domain adaptation} where $|\mathcal{C}_s|=15$, $|\mathcal{C}_t|=60$, $|\mathcal{C}_s \cap \mathcal{C}_t| =10$. The second highest H score is underlined. \textbf{SF} indicates whether source-free.\vspace{0mm}}\label{tab:home_uni}
\scalebox{0.99}{
\begin{tabular}{c|c|cccccccccccc|c}
\hline
\multirow{1}{*}{}&\textbf{SF}
     & A2C&A2P&A2R&C2A&C2P&C2R&P2A&P2C&P2R&R2A&R2C&R2P   & {\textbf{Avg}}  \\\hline
     OSBP~\cite{saito2018open}&\xmark & 39.6 & 45.1 & 46.2 & 45.7 & 45.2 & 46.8 & 45.3 & 40.5 & 45.8 & 45.1 & 41.6 & 46.9 & 44.5 \\
UAN~\cite{you2019universal}&\xmark          & 51.6 & 51.7 & 54.3 & 61.7 & 57.6 & 61.9 & 50.4 & 47.6 & 61.5 & 62.9 & 52.6 & 65.2 & 56.6\\
CMU~\cite{fu2020learning}&\xmark          & 56.0&56.9&59.1&66.9& 64.2&67.8& 54.7& 51.0& 66.3& 68.2&57.8 &69.7&61.6\\
DCC~\cite{li2021domain}&\xmark   &58.0&54.1&58.0&\textbf{74.6}&{70.6}&{77.5}&64.3&\textbf{73.6}&74.9&{81.0}&\textbf{75.1}&{80.4} &70.2       \\
\SQ{DANCE}~\cite{saito2020universal}&\xmark         &-&-&-&-&-&-&-&-&-&-&-&-   & {{49.2}} \\
OVANet~\cite{saito2021ovanet}&\xmark         & {62.8} & {75.6} & {78.6} & {70.7} &{68.8} & {75.0} & {71.3} & {58.6} & {80.5} & {76.1} &{64.1} & {78.9} & \underline{{71.8}} \\
\hline
\textbf{\textit{OneRing}-S}    & &    55.7&72.4&79.6&64.6&65.3&74.6&65.9&51.5&77.9&72.1&57.8&75.0&67.7\\
\textbf{\textit{OneRing}}&\cmark&  {63.3}&  72.4&  {81.0}&  68.8&  67.2&  74.6&  {73.3}&  60.8&  {80.9}&  78.1&  63.9&  76.7&  \underline{71.8} \\
\textbf{\textit{OneRing}+}&\cmark&    \textbf{69.5}&    \textbf{81.4}&    \textbf{87.9}&    73.2&    \textbf{77.9}&    \textbf{82.4}&    \textbf{81.5}&   68.6&    \textbf{88.1}&    \textbf{81.1}&    70.5&    \textbf{85.7}&    \textbf{79.0}\\
\hline
\end{tabular}}
\end{center}
\vspace{0mm}
\end{minipage}
\vspace{0mm}
\end{table*}

\begin{table*}[t]
\addtolength{\tabcolsep}{8pt}
\begin{center}\caption{\textbf{H-score} (\%) on \textbf{DomainNet} using ResNet-50 as backbone. \textbf{Open-partial domain adaptation} where $|\mathcal{C}_s|=200$, $|\mathcal{C}_t|=295$, $|\mathcal{C}_s \cap \mathcal{C}_t| =150$. The second highest H score is underlined. \textbf{SF} indicates whether source-free.\vspace{0mm}}\label{tab:domainnet_opda}
\scalebox{0.99}{
\begin{tabular}{c|c|ccccccc}
\hline
\multirow{1}{*}{Method}&\multirow{1}{*}{\textbf{SF}}& 
     P2R  & R2P  & P2S  & S2P  & R2S  & S2R & \textbf{Avg}  \\\hline
OSBP~\cite{saito2018open} &\xmark&33.6& 33.0& 30.6& 30.5& 30.6& 33.7&32.0 \\
DANCE~\cite{saito2020universal}&\xmark &21.0&47.3&37.0&27.7&\textbf{{46.7}}&21.0&33.5  \\
UAN~\cite{you2019universal}&\xmark  &41.9& 43.6&39.1&38.9& 38.7& 43.7 &41.0 \\
CMU~\cite{fu2020learning}&\xmark   & 50.8 & \textbf{{52.2}} &45.1 & 44.8 & 45.6 & 51.0 & 48.3  \\
DCC~\cite{li2021domain}&\xmark&{56.9} & 50.3&43.7&44.9&43.3& 56.2& 49.2 \\

OVANet~\cite{saito2021ovanet}&\xmark  & 56.0 & 51.7 & {\textbf{47.1}} & {47.4} & {44.9} & {57.2} &  \underline{50.7} \\
\hline

\textbf{\textit{OneRing}-S}&  & \textbf{59.1}&	42.9&	43.8&	35.5&	39.5&	52.9&	45.6   \\
\textbf{\textit{OneRing}}&\cmark  &   57.9&	{52.0}&	46.5&	\textbf{49.6}&	44.1&	\textbf{57.8}&	\textbf{51.3} \\

  \hline
\end{tabular}
}
\end{center}
\vspace{0mm}
\end{table*}

\subsection{Model Details and Evaluation} 
To ensure fair comparison with previous methods, our method is based on the original code released by OPDA method OVANet~\cite{saito2021ovanet} (modified for SF-OPDA). 

\textbf{Training details} 
For SF-OPDA, after finishing source training with Eq.~\ref{eq:ring_src}, we will adapt the source pretrained model to target domain without using source data. Only on the very large DomainNet under SF-OPDA setting we found that our method had difficulties converging (low accuracy on training labeled source data due to the existence of the second CE loss). Therefore, we applied a two-phase training on the source data. In the first phase, we train with the standard CE loss. Then after convergence, we add the second CE loss (multiplied by 1 ((P2R)) or 0.2 (others) to ensure little influence on the accuracy of labeled training data) for a few epochs. For all experiments under SF-OPDA setting, the \textit{OneRing} classifier is fixed during target adaptation. When augmented with AaD~\cite{yang2022attracting}, we set the hyperparameter $K$ in $\mathcal{L}_{dis}$ same as AaD, and $\beta$ in $\mathcal{L}_{div}$ as 1.


\begin{figure}[!tpb]
\centering
    \includegraphics[width=0.99\linewidth]{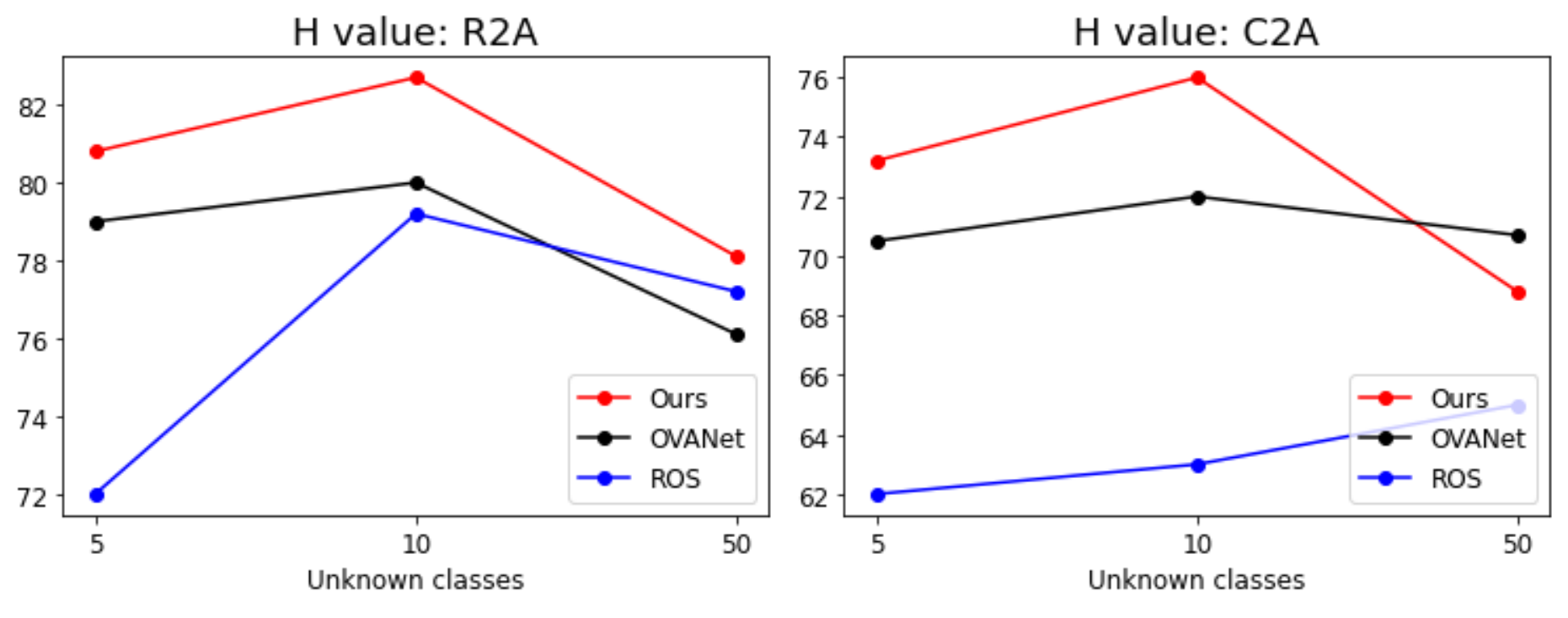}
    \caption{\textbf{H} value of open-partial domain adaptation on Office-Home. We vary the number of unknown classes as shown in the x axis. Here 'ours' denotes OneRing without being augmented with AaD, OVANet and ROS demand source data. 
    \vspace{0mm}}
    \label{fig:unk_num}
\end{figure}

In the experiment, we will mainly report the harmonic mean, as all previous methods did, and also the average per-class accuracy over all categories (OS) on Office-31. Note for SF-OPDA, the model is expected to have high performance on both known and unknown accuracy, which should result in a high harmonic mean (H). As pointed out by ROS~\cite{bucci2020effectiveness}, OS is not a reasonable evaluation metric and can be quite high even when \textit{UNK} is 0, since \textit{OS}$=\frac{|\mathcal{C}_s|}{|\mathcal{C}_s| + 1} \times$\textit{OS$^*$}$+ \frac{1}{|\mathcal{C}_s| + 1} \times$\textit{UNK}. In the following tables, we will denote our model trained with only source data as \textbf{\textit{OneRing}-S}, model after target adaptation as \textbf{\textit{OneRing}}, and model augmented with AaD after target adaptation as \textbf{\textit{One Ring+}}.

\begin{figure*}[ptb]
\centering
    \includegraphics[width=0.99\linewidth]{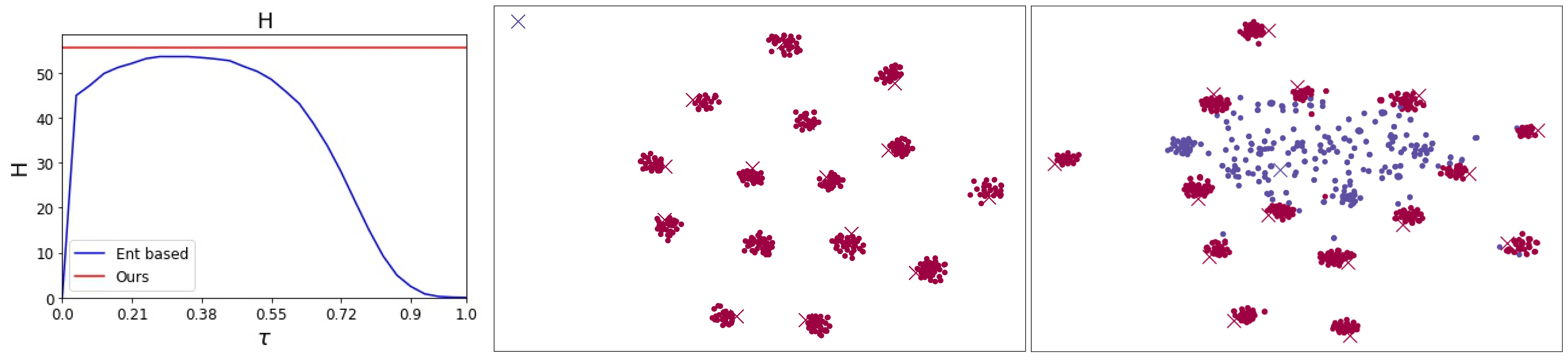}
    \vspace{0mm}
    \caption{(\textbf{Left}) H value of our source model and entropy based rejection on A2C of Office-Home. t-SNE visualization of features with either only source known categories (\textbf{Middle}) or also with 10 source extra unknown categories (\textbf{Right}) from source model on \textit{Artistic} of Office-Home, where the cross is the class prototype. The red denotes known classes while other for unknown class.\vspace{0mm}}
    \label{fig:ent_tsne}
\vspace{0mm}
\end{figure*}


\subsection{Quantitative results}

In Table.~\ref{tab:office31_uni}-\ref{tab:domainnet_opda}, we show the results under open-partial DA setting where \textbf{SF} column indicates whether source-free. Note that our method does not need source data during target adaptation. As shown in the tables, our source model (\textit{One Ring-S}) already achieves decent H performance. The simple {\textit{OneRing}} with only entropy minimization already outperforms all other methods on all 4 benchmarks, adding AaD~\cite{yang2022attracting} into method as shown in Eq.~\ref{eq:final} ({\textit{OneRing}+}) can further improve the results significantly, leading to 0.5\%, 5.4\% and 7.2\% improvement on Office-31, VisDA and Office-Home respectively, and it surpasses the current state-of-the-art OVANet by 2.5\%, 13\% and 7.2\% on these 3 benchmarks respectively. We also show the detailed results of OS*, UNK and H in Table.~\ref{aba_office-home}.

\begin{table}[tbp]
\centering
\caption{Ablation study (R2C of Office-Home) on the proposed weight in the weighted entropy minimization. {Results of OVANet are from our running based on their official code.}} \label{tab10}
\begin{tabular}{c|c|c|c|c}
\hline
\textbf{ R2C }              & \textbf{ OS*} & \textbf{ UNK } & \textbf{ OS } & \textbf{ H }  \\\hline
OVANet~\cite{saito2021ovanet}                      & 55.1                   & 70.0           & 56.5          & 61.7          \\\hline
\textbf{OneRing} w/o weight in Eq.2          & 19.2                   & 97.8           & 26.3          & 32.1          \\
\textbf{OneRing} & 57.8                   & 71.6           & 59.1          & 63.9          \\
\textbf{OneRing+}          & 61.5                   & 82.7           & 63.4          & 70.5    \\     \hline
\end{tabular}
\end{table}

\begin{table*}[tbp]
\centering
\addtolength{\tabcolsep}{-3pt}
\caption{Accuracy (\%) on open-partial DA. \textbf{Results are from one random run.}}\label{tab:home_all}
\resizebox{\textwidth}{!}{
\begin{tabular}{l@{~~~~~~~}c ccc ccc ccc ccc ccc ccc ccc}
\hline

             \multicolumn{15}{c}{~~~~~~~~~~~~~~~~\textbf{Office-Home}}        \\
       
\hline
   & \multicolumn{3}{c|}{Ar $\rightarrow$ Cl } & \multicolumn{3}{c|}{Ar $\rightarrow$ Pr } & \multicolumn{3}{c|}{Ar $\rightarrow$ Rw }  & 
 
 \multicolumn{3}{c|}{Cl$\rightarrow$ Ar } & \multicolumn{3}{c|}{Cl $\rightarrow$ Pr } & \multicolumn{3}{c}{Cl $\rightarrow$ Rw }  \\
       &   OS* & UNK & \multicolumn{1}{c|}{\textbf{\underline{H}}} &  OS* & UNK &  \multicolumn{1}{c|}{\textbf{\underline{H}}} &  OS* & UNK &  \multicolumn{1}{c|}{\textbf{\underline{H}}} &  OS* & UNK &  \multicolumn{1}{c|}{\textbf{\underline{H}}} &  OS* & UNK &  \multicolumn{1}{c|}{\textbf{\underline{H}}} &  OS* & UNK &  \multicolumn{1}{c}{\textbf{\underline{H}}} \\
\hline

\multicolumn{1}{c}{\textbf{OneRing-S}} &  42.9&79.3&55.7 & 75.7&69.5&72.4&91.7&70.3&79.6& 52.9&82.1&64.4&60.0&71.7&65.3& 75.2&74.0&74.6    \\

\multicolumn{1}{c}{\textbf{OneRing}}&  54.1& 73.9& 62.5 & 78.5& 69.8& 73.9 &  93.3& 72.5& 81.6 &  65.9& 73.0& 69.3 & 67.5& 66.1& 66.8 &  80.0& 69.0& 74.1   \\

\multicolumn{1}{c}{\textbf{OneRing+}} & 58.5& 84.4& 69.1 &  78.3&84.8&81.4& 92.6& 84.4& 88.3& 62.7& 88.2& 73.3 &  72.1& 86.3& 78.6 &  80.4& 86.0& 83.1    \\

\hline\hline
 & \multicolumn{3}{c|}{Pr $\rightarrow$ Ar } & \multicolumn{3}{c|}{Pr $\rightarrow$ Cl } & \multicolumn{3}{c|}{Pr $\rightarrow$ Rw }  & 
 
 \multicolumn{3}{c|}{Rw$\rightarrow$ Ar } & \multicolumn{3}{c|}{Rw $\rightarrow$ Cl } & \multicolumn{3}{c}{Rw $\rightarrow$ Pr }& \multicolumn{3}{c}{\textbf{Avg} } \\ 
  
          & OS* & UNK & \multicolumn{1}{c|}{\textbf{\underline{H}}} &  OS* & UNK &  \multicolumn{1}{c|}{\textbf{\underline{H}}} &  OS* & UNK &  \multicolumn{1}{c|}{\textbf{\underline{H}}}&  OS* & UNK &  \multicolumn{1}{c|}{\textbf{\underline{H}}}  & OS* & UNK &  \multicolumn{1}{c|}{\textbf{\underline{H}}}  & OS* & UNK &  \multicolumn{1}{c|}{\textbf{\underline{H}}}& OS* & UNK &  \multicolumn{1}{c|}{\textbf{\underline{H}}}  \\
          
\hline
 \multicolumn{1}{c}{\textbf{OneRing-S}}    & 55.9&80.2&65.9 & 38.6&77.1&51.5 & 86.9&70.5&77.9 & 70.4&73.9&72.1 & 46.6&76.0&57.8 & 82.6&68.7&75.0 & 65.0& 74.4 &67.7  \\

\multicolumn{1}{c}{\textbf{OneRing}}  &   73.1&73.2&73.1 &52.8&70.2&60.3 &91.6&73.4&81.5 & 77.9&78.1&78.0& 57.7&70.2&63.4& 88.2&70.3&78.2 &  73.4 & 71.6  & 71.9\\

\multicolumn{1}{c}{\textbf{OneRing+}}   & 77.4&86.7&82.2 & 58.3&82.3&68.3 & 92.2&84.7&88.3&76.5&86.2&81.1 &61.5&82.7&70.5 & 86.9&85.4&86.1  &  \textbf{74.8} &  \textbf{85.2} & \textbf{79.2} \\
\hline
\end{tabular} \label{aba_office-home}
}
\end{table*}

\begin{table*}[!h]
\centering
\addtolength{\tabcolsep}{4pt}
\caption{\textbf{H-score} (\%) on \textbf{Office-Home} dataset using ResNet-50 as backbone. \textbf{Open-partial domain adaptation} where $|\mathcal{C}_s|=15$, $|\mathcal{C}_t|=60$, $|\mathcal{C}_s \cap \mathcal{C}_t| =10$. The second highest H score is underlined. \textbf{SF} indicates whether source-free. {* indicates using predictions over the \textit{whole dataset} instead of mini-batch in Eq.~\ref{eq:ring_tar}.}\vspace{0mm}}\label{tab:home_uni_all_batch}
\scalebox{0.99}{
\begin{tabular}{c|cccccccccccc|c}
\hline
\multirow{1}{*}{}
     & A2C&A2P&A2R&C2A&C2P&C2R&P2A&P2C&P2R&R2A&R2C&R2P   & {\textbf{Avg}}  \\\hline
OVANet~\cite{saito2021ovanet}      & {62.8} & {75.6} & {78.6} & {70.7} &{68.8} & {75.0} & {71.3} & {58.6} & {80.5} & {76.1} &{64.1} & {78.9} & {{71.8}} \\
\hline
\textbf{\textit{OneRing}-S}   &    55.7&72.4&79.6&64.6&65.3&74.6&65.9&51.5&77.9&72.1&57.8&75.0&67.7\\
\textbf{\textit{OneRing}}&  {63.3}&  72.4&  {81.0}&  68.8&  67.2&  74.6&  {73.3}&  60.8&  {80.9}&  78.1&  63.9&  76.7&  {71.8} \\
{\textbf{\textit{OneRing*}}}&  {60.9}&  72.1&  {80.9}&  67.7&  66.0&  73.7&  {73.1}&  60.4&  {81.4}&  77.7&  63.4&  78.2&  {71.3} \\
\textbf{\textit{OneRing}+}&    {69.5}&    {81.4}&    {87.9}&    73.2&    {77.9}&    {82.4}&    {81.5}&   68.6&    {88.1}&    {81.1}&    70.5&    {85.7}&    {79.0}\\
{\textbf{\textit{OneRing*}+}}&    {70.1}&    {82.5}&    {88.9}&  75.1   &    {80.1}&    {83.0}&    {82.5}&   64.6&    {89.3}&    {81.0}&  66.4   &    {86.0}&    \textbf{79.1}\\
\hline
\end{tabular}}

\vspace{0mm}
\end{table*}

\subsection{Analysis}

\textbf{Compare One Ring with entropy based unknown rejection.} We also show the results with entropy based unknown rejection, where a sample is predicted as unknown if the entropy (maximal normalized) of the prediction (\textit{with normal classifier head}) is higher than a manually set threshold. Fig.~\ref{fig:ent_tsne} (\textit{left}) shows the H value of \textit{source pretrained model} on A2C task of Office-Home under open-partial DA setting, where the \textit{x axis} denotes the threshold. Our source model gets better results without any extra effort.



\textbf{Visualization of features and class prototypes.} In Fig.~\ref{fig:ent_tsne} (\textit{Middle}), we visualize the source features and class prototypes (weights of \textit{OneRing} classifier) from \textit{source model} with t-SNE. The prototype of the unknown category is in the corner with no source features around it. In Fig.~\ref{fig:ent_tsne} (\textit{Right}), we further visualize 10 extra unknown classes. It shows that those features of unknown categories will not cluster around any of the known classes, but they are close to the unknown prototype. This implies that the \textit{OneRing} model can efficiently distinguish known and unknown categories.

\textbf{Importance of weight in entropy minimization.} We ablate the weights in entropy minimization in Eq.~\ref{eq:ring_tar}. If removing weights, the \textit{OS*}, \textit{UNK} and \textit{H} on R2C (Office-Home) will decrease. In Table.~\ref{tab10}, we report OS*, UNK, OS and H on VisDA under open-partial DA, we outperform OVANet on the metrics of both OS and H. It shows the deployed weights are important and effective to balance the two terms in Eq.~\ref{eq:ring_tar}.


\textbf{Robustness to amount of unknown categories.} In Fig.~\ref{fig:unk_num}, we compare our source-free \textit{OneRing} (without being augmented with AaD) to ROS~\cite{bucci2020effectiveness} and OVANet~\cite{saito2021ovanet} under OPDA setting with different amount of unknown categories from target domain. The results show that our method is robust to the amount of unknown categories.

\textbf{Known/unknown ratio estimation through mini-batch or whole dataset.} In Eq. 2, we have two choice to estimate the known/unknown ratio, which will be utilized to balance the 2 entropy terms. In Table.~\ref{tab:home_uni_all_batch}, we show that these 2 different manners lead to almost the same results. Though there may exist some imbalance mini-batches which only contain few samples predicted as known or unknown, the results imply that the known/unknown ratio estimated by the mini-batch is enough to achieve decent performance. Note the Office-Home here is not a well balance (amount of samples per category) dataset, and also in the target domain the unknown categories (50) are much more than known (10).

\section{Conclusion}
In this paper, we first introduce a simple method with the proposed \textit{OneRing} classifier head, it possesses strong ability to detect unknown categories from target data even no matter without or with domain shift after training with two simple cross entropy losses. Then, we further adapt the model to the target domain which contains unknown categories, with only weighted entropy minimization and no access to source data. In the experiment, we show that our method achieves good performance on source-free open-partial domain adaptation, which proves the effectiveness of our method.

\ifCLASSOPTIONcompsoc
  \section*{Acknowledgments}
\else
  \section*{Acknowledgment}
\fi

We acknowledge the support from Huawei Kirin Solution. We further acknowledge the Spanish projects PID2019-104174GB-I00 and RTI2018-102285-A-I0, and the CERCA Program of the Generalitat de Catalunya. We thank Qiuyue Yang for drawing the excellent figure.

\ifCLASSOPTIONcaptionsoff
  \newpage
\fi

{\small
\bibliography{longstrings,iclr2023_conference}

\begin{thebibliography}{10}

\bibitem{bucci2020effectiveness}
Silvia Bucci, Mohammad~Reza Loghmani, and Tatiana Tommasi.
\newblock On the effectiveness of image rotation for open set domain
  adaptation.
\newblock In {\em ECCV}, pages 422--438. Springer, 2020.

\bibitem{cao2018partial}
Zhangjie Cao, Lijia Ma, Mingsheng Long, and Jianmin Wang.
\newblock Partial adversarial domain adaptation.
\newblock In {\em ECCV}, pages 135--150, 2018.

\bibitem{cao2019learning}
Zhangjie Cao, Kaichao You, Mingsheng Long, Jianmin Wang, and Qiang Yang.
\newblock Learning to transfer examples for partial domain adaptation.
\newblock In {\em CVPR}, pages 2985--2994, 2019.

\bibitem{changunified}
Wanxing Chang, Ye~Shi, Hoang~Duong Tuan, and Jingya Wang.
\newblock Unified optimal transport framework for universal domain adaptation.
\newblock In {\em Advances in Neural Information Processing Systems}, 2022.

\bibitem{chen2021adversarial}
Guangyao Chen, Peixi Peng, Xiangqian Wang, and Yonghong Tian.
\newblock Adversarial reciprocal points learning for open set recognition.
\newblock {\em IEEE TPAMI}, 2021.

\bibitem{Chen_2020_ECCV}
Guangyao Chen, Limeng Qiao, Yemin Shi, Peixi Peng, Jia Li, Tiejun Huang,
  Shiliang Pu, and Yonghong Tian.
\newblock Learning open set network with discriminative reciprocal points.
\newblock In {\em ECCV}, 2020.

\bibitem{cicek2019unsupervised}
Safa Cicek and Stefano Soatto.
\newblock Unsupervised domain adaptation via regularized conditional alignment.
\newblock In {\em ICCV}, pages 1416--1425, 2019.

\bibitem{cui2020towards}
Shuhao Cui, Shuhui Wang, Junbao Zhuo, Liang Li, Qingming Huang, and Qi~Tian.
\newblock Towards discriminability and diversity: Batch nuclear-norm
  maximization under label insufficient situations.
\newblock {\em CVPR}, 2020.

\bibitem{deng2019cluster}
Zhijie Deng, Yucen Luo, and Jun Zhu.
\newblock Cluster alignment with a teacher for unsupervised domain adaptation.
\newblock In {\em ICCV}, pages 9944--9953, 2019.

\bibitem{feng2019attract}
Qianyu Feng, Guoliang Kang, Hehe Fan, and Yi~Yang.
\newblock Attract or distract: Exploit the margin of open set.
\newblock In {\em ICCV}, pages 7990--7999, 2019.

\bibitem{feng2021open}
Zeyu Feng, Chang Xu, and Dacheng Tao.
\newblock Open-set hypothesis transfer with semantic consistency.
\newblock {\em IEEE TIP}, 30:6473--6484, 2021.

\bibitem{fu2020learning}
Bo~Fu, Zhangjie Cao, Mingsheng Long, and Jianmin Wang.
\newblock Learning to detect open classes for universal domain adaptation.
\newblock In {\em ECCV}, pages 567--583. Springer, 2020.

\bibitem{ganin2016domain}
Yaroslav Ganin, Evgeniya Ustinova, Hana Ajakan, Pascal Germain, Hugo
  Larochelle, Fran{\c{c}}ois Laviolette, Mario Marchand, and Victor Lempitsky.
\newblock Domain-adversarial training of neural networks.
\newblock {\em JMLR}, 17(1):2096--2030, 2016.

\bibitem{Ge2017Generative}
Zongyuan Ge, Sergey Demyanov, and Rahil Garnavi.
\newblock Generative openmax for multi-class open set classification.
\newblock In {\em BMVC}, 2017.

\bibitem{gulrajani2020search}
Ishaan Gulrajani and David Lopez-Paz.
\newblock In search of lost domain generalization.
\newblock {\em ICLR}, 2021.

\bibitem{huang2021model}
Jiaxing Huang, Dayan Guan, Aoran Xiao, and Shijian Lu.
\newblock Model adaptation: Historical contrastive learning for unsupervised
  domain adaptation without source data.
\newblock {\em NeurIPS}, 34, 2021.

\bibitem{jing2021towards}
Taotao Jing, Hongfu Liu, and Zhengming Ding.
\newblock Towards novel target discovery through open-set domain adaptation.
\newblock In {\em ICCV}, pages 9322--9331, 2021.

\bibitem{kong2021}
Shu Kong and Deva Ramanan.
\newblock Opengan: Open-set recognition via open data generation.
\newblock {\em ICCV}, 2021.

\bibitem{kundusubsidiary}
Jogendra~Nath Kundu, Suvaansh Bhambri, Akshay~Ravindra Kulkarni, Hiran Sarkar,
  Varun Jampani, and Venkatesh~Babu Radhakrishnan.
\newblock Subsidiary prototype alignment for universal domain adaptation.
\newblock In {\em Advances in Neural Information Processing Systems}, 2022.

\bibitem{kundu2020universal}
Jogendra~Nath Kundu, Naveen Venkat, and R~Venkatesh Babu.
\newblock Universal source-free domain adaptation.
\newblock {\em CVPR}, 2020.

\bibitem{kundu2020towards}
Jogendra~Nath Kundu, Naveen Venkat, Ambareesh Revanur, R~Venkatesh Babu, et~al.
\newblock Towards inheritable models for open-set domain adaptation.
\newblock In {\em CVPR}, pages 12376--12385, 2020.

\bibitem{lee2019sliced}
Chen-Yu Lee, Tanmay Batra, Mohammad~Haris Baig, and Daniel Ulbricht.
\newblock Sliced wasserstein discrepancy for unsupervised domain adaptation.
\newblock In {\em CVPR}, pages 10285--10295, 2019.

\bibitem{li2021domain}
Guangrui Li, Guoliang Kang, Yi~Zhu, Yunchao Wei, and Yi~Yang.
\newblock Domain consensus clustering for universal domain adaptation.
\newblock In {\em CVPR}, pages 9757--9766, 2021.

\bibitem{li2020model}
Rui Li, Qianfen Jiao, Wenming Cao, Hau-San Wong, and Si~Wu.
\newblock Model adaptation: Unsupervised domain adaptation without source data.
\newblock In {\em CVPR}, pages 9641--9650, 2020.

\bibitem{liang2020we}
Jian Liang, Dapeng Hu, and Jiashi Feng.
\newblock Do we really need to access the source data? source hypothesis
  transfer for unsupervised domain adaptation.
\newblock {\em ICML}, 2020.

\bibitem{liang2021domain}
Jian Liang, Dapeng Hu, and Jiashi Feng.
\newblock Domain adaptation with auxiliary target domain-oriented classifier.
\newblock In {\em CVPR}, pages 16632--16642, 2021.

\bibitem{liang2021umad}
Jian Liang, Dapeng Hu, Jiashi Feng, and Ran He.
\newblock Umad: Universal model adaptation under domain and category shift.
\newblock {\em arXiv preprint arXiv:2112.08553}, 2021.

\bibitem{liang2020balanced}
Jian Liang, Yunbo Wang, Dapeng Hu, Ran He, and Jiashi Feng.
\newblock A balanced and uncertainty-aware approach for partial domain
  adaptation.
\newblock In {\em ECCV}, pages 123--140. Springer, 2020.

\bibitem{liu2019separate}
Hong Liu, Zhangjie Cao, Mingsheng Long, Jianmin Wang, and Qiang Yang.
\newblock Separate to adapt: Open set domain adaptation via progressive
  separation.
\newblock In {\em CVPR}, pages 2927--2936, 2019.

\bibitem{liu2021cycle}
Hong Liu, Jianmin Wang, and Mingsheng Long.
\newblock Cycle self-training for domain adaptation.
\newblock In {\em NeurIPS}, 2021.

\bibitem{long2018transferable}
Mingsheng Long, Yue Cao, Zhangjie Cao, Jianmin Wang, and Michael~I Jordan.
\newblock Transferable representation learning with deep adaptation networks.
\newblock {\em TPAMI}, 41(12):3071--3085, 2018.

\bibitem{long2015learning}
Mingsheng Long, Yue Cao, Jianmin Wang, and Michael~I Jordan.
\newblock Learning transferable features with deep adaptation networks.
\newblock {\em ICML}, 2015.

\bibitem{long2018conditional}
Mingsheng Long, Zhangjie Cao, Jianmin Wang, and Michael~I Jordan.
\newblock Conditional adversarial domain adaptation.
\newblock In {\em NIPS}, pages 1647--1657, 2018.

\bibitem{long2016unsupervised}
Mingsheng Long, Han Zhu, Jianmin Wang, and Michael~I Jordan.
\newblock Unsupervised domain adaptation with residual transfer networks.
\newblock In {\em NIPS}, pages 136--144, 2016.

\bibitem{lu2020stochastic}
Zhihe Lu, Yongxin Yang, Xiatian Zhu, Cong Liu, Yi-Zhe Song, and Tao Xiang.
\newblock Stochastic classifiers for unsupervised domain adaptation.
\newblock In {\em CVPR}, pages 9111--9120, 2020.

\bibitem{Neal_2018_ECCV}
Lawrence Neal, Matthew Olson, Xiaoli Fern, Weng-Keen Wong, and Fuxin Li.
\newblock Open set learning with counterfactual images.
\newblock In {\em ECCV}, 2018.

\bibitem{pan2020exploring}
Yingwei Pan, Ting Yao, Yehao Li, Chong-Wah Ngo, and Tao Mei.
\newblock Exploring category-agnostic clusters for open-set domain adaptation.
\newblock In {\em CVPR}, pages 13867--13875, 2020.

\bibitem{panareda2017open}
Pau Panareda~Busto and Juergen Gall.
\newblock Open set domain adaptation.
\newblock In {\em ICCV}, pages 754--763, 2017.

\bibitem{peng2019moment}
Xingchao Peng, Qinxun Bai, Xide Xia, Zijun Huang, Kate Saenko, and Bo~Wang.
\newblock Moment matching for multi-source domain adaptation.
\newblock In {\em ICCV}, pages 1406--1415, 2019.

\bibitem{peng2017visda}
Xingchao Peng, Ben Usman, Neela Kaushik, Judy Hoffman, Dequan Wang, and Kate
  Saenko.
\newblock Visda: The visual domain adaptation challenge.
\newblock {\em arXiv preprint arXiv:1710.06924}, 2017.

\bibitem{robey2021model}
Alexander Robey, George~J Pappas, and Hamed Hassani.
\newblock Model-based domain generalization.
\newblock {\em NeurIPS}, 34:20210--20229, 2021.

\bibitem{saito2020universal}
Kuniaki Saito, Donghyun Kim, Stan Sclaroff, and Kate Saenko.
\newblock Universal domain adaptation through self supervision.
\newblock {\em NeurIPS}, 33, 2020.

\bibitem{saito2021ovanet}
Kuniaki Saito and Kate Saenko.
\newblock Ovanet: One-vs-all network for universal domain adaptation.
\newblock In {\em ICCV}, pages 9000--9009, 2021.

\bibitem{saito2018maximum}
Kuniaki Saito, Kohei Watanabe, Yoshitaka Ushiku, and Tatsuya Harada.
\newblock Maximum classifier discrepancy for unsupervised domain adaptation.
\newblock In {\em CVPR}, pages 3723--3732, 2018.

\bibitem{saito2018open}
Kuniaki Saito, Shohei Yamamoto, Yoshitaka Ushiku, and Tatsuya Harada.
\newblock Open set domain adaptation by backpropagation.
\newblock In {\em ECCV}, pages 153--168, 2018.

\bibitem{shi2021gradient}
Yuge Shi, Jeffrey Seely, Philip~HS Torr, N~Siddharth, Awni Hannun, Nicolas
  Usunier, and Gabriel Synnaeve.
\newblock Gradient matching for domain generalization.
\newblock {\em ICLR}, 2022.

\bibitem{shu2018dirt}
Rui Shu, Hung~H Bui, Hirokazu Narui, and Stefano Ermon.
\newblock A dirt-t approach to unsupervised domain adaptation.
\newblock {\em ICLR}, 2018.

\bibitem{shu2021open}
Yang Shu, Zhangjie Cao, Chenyu Wang, Jianmin Wang, and Mingsheng Long.
\newblock Open domain generalization with domain-augmented meta-learning.
\newblock In {\em CVPR}, pages 9624--9633, 2021.

\bibitem{Shu2020podn}
Yu~Shu, Yemin Shi, Yaowei Wang, Tiejun Huang, and Yonghong Tian.
\newblock P-odn: Prototype-based open deep network for open set recognition.
\newblock {\em Scientific Reports}, 2020.

\bibitem{sun2016return}
Baochen Sun, Jiashi Feng, and Kate Saenko.
\newblock Return of frustratingly easy domain adaptation.
\newblock In {\em AAAI}, 2016.

\bibitem{Sun_2020_CVPR}
Xin Sun, Zhenning Yang, Chi Zhang, Guohao Peng, and Keck-Voon Ling.
\newblock Conditional gaussian distribution learning for open set recognition.
\newblock In {\em CVPR}, 2020.

\bibitem{tang2020unsupervised}
Hui Tang, Ke~Chen, and Kui Jia.
\newblock Unsupervised domain adaptation via structurally regularized deep
  clustering.
\newblock In {\em CVPR}, pages 8725--8735, 2020.

\bibitem{tzeng2017adversarial}
Eric Tzeng, Judy Hoffman, Kate Saenko, and Trevor Darrell.
\newblock Adversarial discriminative domain adaptation.
\newblock In {\em CVPR}, pages 7167--7176, 2017.

\bibitem{tzeng2014deep}
Eric Tzeng, Judy Hoffman, Ning Zhang, Kate Saenko, and Trevor Darrell.
\newblock Deep domain confusion: Maximizing for domain invariance.
\newblock {\em arXiv preprint arXiv:1412.3474}, 2014.

\bibitem{vaze2022openset}
Sagar Vaze, Kai Han, Andrea Vedaldi, and Andrew Zisserman.
\newblock Open-set recognition: A good closed-set classifier is all you need.
\newblock In {\em ICLR}, 2022.

\bibitem{vedantam2021empirical}
Ramakrishna Vedantam, David Lopez-Paz, and David~J Schwab.
\newblock An empirical investigation of domain generalization with empirical
  risk minimizers.
\newblock {\em NeurIPS}, 34, 2021.

\bibitem{wang2021learning}
Zijian Wang, Yadan Luo, Ruihong Qiu, Zi~Huang, and Mahsa Baktashmotlagh.
\newblock Learning to diversify for single domain generalization.
\newblock In {\em ICCV}, pages 834--843, 2021.

\bibitem{wu2018unsupervised}
Zhirong Wu, Yuanjun Xiong, Stella~X Yu, and Dahua Lin.
\newblock Unsupervised feature learning via non-parametric instance
  discrimination.
\newblock In {\em CVPR}, pages 3733--3742, 2018.

\bibitem{xia2021adaptive}
Haifeng Xia, Handong Zhao, and Zhengming Ding.
\newblock Adaptive adversarial network for source-free domain adaptation.
\newblock In {\em ICCV}, pages 9010--9019, 2021.

\bibitem{yang2021exploiting}
Shiqi Yang, Joost van~de Weijer, Luis Herranz, Shangling Jui, et~al.
\newblock Exploiting the intrinsic neighborhood structure for source-free
  domain adaptation.
\newblock {\em NeurIPS}, 34, 2021.

\bibitem{yang2020unsupervised}
Shiqi Yang, Yaxing Wang, Joost van~de Weijer, Luis Herranz, and Shangling Jui.
\newblock Unsupervised domain adaptation without source data by casting a bait.
\newblock {\em arXiv preprint arXiv:2010.12427}, 2020.

\bibitem{yang2021generalized}
Shiqi Yang, Yaxing Wang, Joost van~de Weijer, Luis Herranz, and Shangling Jui.
\newblock Generalized source-free domain adaptation.
\newblock In {\em ICCV}, pages 8978--8987, 2021.

\bibitem{yang2022attracting}
Shiqi Yang, Yaxing Wang, Kai Wang, Shangling Jui, et~al.
\newblock Attracting and dispersing: A simple approach for source-free domain
  adaptation.
\newblock In {\em NeurIPS}, 2022.

\bibitem{you2019universal}
Kaichao You, Mingsheng Long, Zhangjie Cao, Jianmin Wang, and Michael~I Jordan.
\newblock Universal domain adaptation.
\newblock In {\em CVPR}, pages 2720--2729, 2019.

\bibitem{zhang2020hybrid}
Hongjie Zhang, Ang Li, Jie Guo, and Yanwen Guo.
\newblock Hybrid models for open set recognition.
\newblock In {\em ECCV}, pages 102--117. Springer, 2020.

\bibitem{zhang2019domain}
Yabin Zhang, Hui Tang, Kui Jia, and Mingkui Tan.
\newblock Domain-symmetric networks for adversarial domain adaptation.
\newblock In {\em CVPR}, pages 5031--5040, 2019.

\bibitem{zhou2021learning}
Da-Wei Zhou, Han-Jia Ye, and De-Chuan Zhan.
\newblock Learning placeholders for open-set recognition.
\newblock In {\em CVPR}, pages 4401--4410, 2021.

\bibitem{zhu2022crossmatch}
Ronghang Zhu and Sheng Li.
\newblock Crossmatch: Cross-classifier consistency regularization for open-set
  single domain generalization.
\newblock In {\em ICLR}, 2022.

\end{thebibliography}
\bibliographystyle{plain}
}

\begin{IEEEbiographynophoto}{Shiqi Yang}
joined Learning and Machine Perception (LAMP) team in 2019.10 as a Ph.D. student advised by Dr. Joost van de Weijer in Computer Vision Center of Autonomous University of Barcelona, Spain. He received master degree in Huazhong University of Science and Technology, China, and once worked as a research associate in Kyoto University, Japan. His research interest focuses on
how to efficiently adapt the pretrained model to real
world environment under domain and category shift,
including source-free/continual/open-set/universal domain adaptation.
\end{IEEEbiographynophoto}

\begin{IEEEbiographynophoto}{Yaxing Wang}
is an associate professor of college of computer science at Nankai University. His research interests include GANs, image-to-image translation, domain adaptation and lifelong learning. Prior to joining NKU I was a postdoc at UAB, CVC, working with Joost van de Weije. I obtained my Ph.D. from Universitat Autònoma de Barcelona, under the supervision of Joost van de Weijer. I also experienced amazing internship at IIAI with Fahad shahbaz khan and Salman Khan.
\end{IEEEbiographynophoto}

\begin{IEEEbiographynophoto}{Kai Wang} is a postdoctoral researcher at Computer Vision Center, UAB. Before he obtained the Ph.D. degree from Computer Vision Center, UAB in 2022 under the supervision of Joost van de Weijer. He received the master degree in image processing from Jilin University in 2017 and the bachelor degree from Jilin University in 2014. His main research interests include continual learning, knowledge distillation, domain adaptation and vision transformers.
\end{IEEEbiographynophoto}

\begin{IEEEbiographynophoto}{Shangling Jui}
is the Chief AI Scientist for Huawei Kirin Chipset Solution. He is an expert in machine learning, deep learning, and artificial intelligence. Previously, he was the President of the SAP China Research Center and the SAP Korea Research Center, responsible for 2400 employees and 150 million USD research and development annual budget. He was also the CTO of Pactera, leading innovation projects based on cloud and big data technologies. He is currently an Expert Reviewer of the Project Committee for China-EU Science and Technology Co-Operation and a Guest Professor of the Software Institute of Beijing University. He has published various books and articles about the Chinese software industry and big data analytics in China, U.K., Australia, and the USA. He has 27 years of working experience in Germany, the USA, and China. He received the Magnolia Award from the Municipal Government of Shanghai, in 2011.
\end{IEEEbiographynophoto}

\begin{IEEEbiographynophoto}{Joost van de Weijer}
received the Ph.D. degree
from the University of Amsterdam, Amsterdam,
Netherlands, in 2005.
He was a Marie Curie Intra-European Fellow with INRIA Rhone-Alpes, France, and from
2008 to 2012, he was a Ramon y Cajal Fellow with
the Universitat Autònoma de Barcelona, Barcelona,
Spain, where he is currently a Senior Scientist
with the Computer Vision Center and leader of the
Learning and Machine Perception (LAMP) Team.
His main research directions are color in computer
vision, continual learning, active learning, and domain adaptation.
\end{IEEEbiographynophoto}





\end{document}